# Research on Aerodynamic Performance Prediction of Airfoils Based on a Fusion Algorithm of Transformer and GAN

MaolinYang, Yaohui Wang, Pingyu Jiang[a]*

[a]*State Key Laboratory of Manufacturing Systems Engineering, Xi'an Jiaotong University, Xi'an, 710054, China*

Corresponding to: pjiang@xjtu.edu.cn

Abstract - Predicting of airfoil aerodynamic performance is a key part of aircraft design optimization. but the traditional methods (such as wind tunnel test and CFD simulation) have the problems of high cost and low efficiency, and the existing data-driven models face the challenges of insufficient accuracy and strong data dependence in multi-objective prediction. Therefore, this study proposes a deep learning model——Deeptrans, based on the fusion of improved Transformer and generative Adversarial network (GAN), which aims to predict the multi-parameter aerodynamic performance of airfoil efficiently. By constructing a large-scale data set and designing a model structure that integrates a Transformer coding-decoding framework and confrontation training, synchronous and high-precision prediction of aerodynamic parameters is realized. Experiments show that the MSE loss of Deeptrans on the verification set is reduced to $5.6 \times 10^{-6}$, and the single-sample prediction time is only 0.0056 seconds, which is nearly 700 times more efficient than the traditional CFD method. Horizontal comparison shows that the prediction accuracy is significantly better than the original Transformer, GAN, and VAE models. This study provides an efficient data-driven solution for airfoil aerodynamic performance prediction and a new idea for deep learning modeling complex flow problems.

# 1 Introduction

The aerodynamic performance of airfoils is one of the critical factors in aircraft design, directly influencing core metrics such as lift, drag, stability, and fuel efficiency. Key parameters for evaluating airfoil performance include the lift coefficient, drag coefficient, pressure drag coefficient, pitching moment coefficient, and the location of the transition point. Accurate prediction of these parameters not only optimizes airfoil design but also shortens the aircraft development cycle and reduces experimental costs. In modern aviation, with increasingly stringent performance requirements for aircraft, higher demands are placed on the precision and efficiency of aerodynamic performance prediction. Therefore, developing efficient and accurate methods for aerodynamic performance prediction holds significant engineering and academic value.

Traditionally, wind tunnel tests and numerical simulations based on Computational Fluid Dynamics (CFD) have been the main methods for obtaining data on aerodynamic performance parameters of airfoils. While wind tunnel testing provides high-precision experimental data, it is costly, time-consuming, and constrained by experimental complexity. CFD simulations, though theoretically capable of modeling complex aerodynamic phenomena, require substantial computational resources and time, particularly when addressing turbulence, transition, and other intricate flow phenomena, often struggling to balance accuracy and efficiency. Numerous surrogate modeling approaches have been explored to reduce costs. Traditional surrogate models, such as polynomial response surfaces (Box & Draper, 1987), Kriging (Sacks et al., 1989), and support vector regression (SVR) (Smola & Schölkopf, 2004), partially mitigate computational expenses but underperform with large training datasets and fail to capture complex nonlinear relationships. To address these limitations, neural network models—including multilayer perceptrons (MLP) (Santos, J. E., et al., 2018), convolutional neural networks (CNN), variational autoencoders (VAE), and generative adversarial networks (GAN) (Du, X.S., et al., 2021)—have been adopted. These models learn the mapping between airfoil geometric parameters and aerodynamic performance, significantly improving prediction accuracy and efficiency. For instance, Sekar et al. (2019) combined CNN-extracted geometric features with MLP to predict flow fields, Du et al. (2021) leveraged B-spline parameterization and GANs for rapid aerodynamic prediction and optimization.

Despite advancements, traditional surrogate models and neural networks still exhibit shortcomings. First, traditional surrogate models struggle with high-dimensional data and nonlinear relationships. Second, existing neural networks (e.g., MLP, CNN) face challenges in balancing accuracy and efficiency for multi-objective prediction tasks. Additionally, most models heavily depend on data quality and quantity, leading to degraded performance under sparse or noisy data conditions. These limitations hinder the practical application of existing methods in engineering. For example, while parametric methods like singular value decomposition (SVD) (Masters, D. A., et al., 2017) and class shape transformation (CST) (Kulfan, B. M., 2008) enable compact airfoil representation, they often fail to capture intricate geometric details of unconventional airfoils. Similarly, image-based CNN approaches (Yilmaz, E., & German, B., 2020). face resolution-dependent accuracy issues and computational bottlenecks when handling large flow fields.

Airfoil parameterization is a critical step in aerodynamic performance prediction, aiming to convert complex geometric shapes into manageable parameters. Common parameterization methods include direct coordinate-based representation, spline curve-based approaches (e.g., B-spline, NURBS), and modal decomposition techniques (e.g., PCA). Each method has trade-offs: coordinate-based representation is intuitive but high-dimensional, while spline-based methods efficiently describe shapes with fewer control points but may lose local details. Recent advances in deep learning have enabled data-driven parameterization methods (Chen et al., 2017), which automatically extract features from large airfoil datasets to achieve more efficient representations.

Selecting appropriate input features is crucial for successful aerodynamic performance prediction. Airfoil coordinates directly describe geometric shapes and serve as the foundation for predictions. Parameters such as thickness distribution (n-value) further simplify geometric descriptions, while the Reynolds number reflects flow state effects. Combining these inputs comprehensively captures geometric and flow characteristics, enhancing prediction accuracy and generalization. Studies demonstrate that deep learning models using these inputs excel in aerodynamic performance prediction (Zhang et al., 2018).

To address these challenges, this study proposes an improved Transformer-based method for aerodynamic performance prediction. The Transformer model, renowned for its sequence modeling capabilities and parallel computation efficiency, has achieved remarkable success in natural language processing and computer vision (Vaswani et al., 2017). This work introduces the Transformer to aerodynamic prediction and integrates it with generative adversarial network (GAN) principles, proposing the Deeptrans model. Deeptrans simultaneously predicts multiple parameters—lift coefficient, drag coefficient, pressure drag coefficient, pitching moment coefficient, and transition point location—significantly improving prediction speed and accuracy.

The structure of this paper is as follows: Chapter 1 introduces the research background, significance, and limitations of existing methods. Chapter 2 reviews the literature on airfoil coordinate datasets, Transformer and GAN algorithms, and deep learning applications in aerodynamic prediction. Chapter 3 details the Deeptrans model framework, dataset construction, and training processes. Chapter 4 validates the model's effectiveness through case studies. Chapter 5 compares Deeptrans with Transformer, VAE, and GAN models. Chapter 6 concludes the research and outlines future directions.

Through this work, we aim to provide a high-efficiency, high-accuracy solution for airfoil aerodynamic performance prediction, offering robust support for aircraft design and optimization.

## 2 Literature Review

### *2.1 Airfoil Coordinates and Aerodynamic Performance Dataset*

Airfoil coordinate datasets constitute the fundamental cornerstone of aerodynamic performance prediction research. Current airfoil geometric characterization methods primarily rely on discrete coordinate systems, with representative public databases including the UIUC Airfoil Coordinates Database, the Airfoil Tools online platform database, and the Profili integrated database. These resources systematically compile geometric topology data of canonical standard airfoil profiles. Notably, geometric coordinate information alone cannot directly facilitate aerodynamic performance prediction modeling, necessitating the coupling with key aerodynamic parameters such as lift and drag coefficients to construct data-driven models. The existing database architecture exhibits dual limitations: First, mainstream datasets exemplified by UIUC and Profili only provide basic geometric coordinates while lacking corresponding aerodynamic parameters, whereas the high-precision predictive capabilities of deep learning models depend on large-scale, high-fidelity datasets encompassing geometry-aerodynamic parameter mapping relationships (Duraisamy et al., 2019). Second, platforms like Airfoil Tools demonstrate significantly delayed updating efficiency for aerodynamic parameters of novel airfoil configurations (e.g., asymmetric topological structures or flap-integrated designs) relative to research demands. This data discontinuity compels researchers to rely on wind tunnel experiments or computational fluid dynamics (CFD) numerical simulations to acquire target parameters, substantially hindering research efficiency. Consequently, establishing an integrated database that combines airfoil geometric features with multi-physics aerodynamic parameters has emerged as a critical breakthrough direction for enhancing the accuracy of data-driven aerodynamic analysis.

*2.2 Deep Learning Model Prediction of Airfoil Aerodynamic Performance*

In recent years, deep learning models have been widely applied to airfoil aerodynamic performance prediction. Early studies primarily utilized multilayer perceptrons (MLP) to predict parameters such as lift coefficient ($C_L$) and drag coefficient ($C_D$) by learning the mapping between airfoil geometric parameters and aerodynamic performance (Thuerey et al., 2020). However, MLPs underperform with high-dimensional input data and struggle to capture complex nonlinear relationships.

To overcome these limitations, researchers have adopted convolutional neural networks (CNN). CNNs extract geometric features of airfoils through convolutional operations, significantly improving prediction accuracy. For example, Zhang et al. proposed a CNN-based model for airfoil aerodynamic performance prediction, capable of simultaneously predicting multiple aerodynamic parameters and outperforming traditional methods across diverse datasets. Additionally, generative adversarial networks (GAN) and variational autoencoders (VAE) have been introduced to this field. GANs leverage adversarial training between generators and discriminators to synthesize high-fidelity aerodynamic performance data, while VAEs enhance robustness to data sparsity and noise by learning latent variable representations (Kingma & Welling, 2013).

Despite these advancements, challenges remain. Existing models often fail to balance accuracy and efficiency in multi-objective prediction tasks and exhibit strong dependence on data quality and quantity. Furthermore, most models lack effective modeling capabilities for sequential data (e.g., airfoil coordinate sequences), limiting their performance with high-dimensional inputs.

Recently, the Transformer model has been introduced to scientific computing and engineering, demonstrating significant potential in fluid dynamics prediction. A notable example is the work by Zhou et al.(Zhou et al., 2023), who developed a Transformer-based neural network specifically designed for predicting flow fields over airfoils. Researchers have also developed multiple Transformer variants to enhance its capabilities. Among these, Transformer-GAN hybrids have garnered attention. By integrating GAN's adversarial training mechanism, such models can generate high-fidelity data and improve the modeling of complex distributions (Zhang et al., 2022). Combining Transformer's sequence modeling strengths with GAN's adversarial training enhances the physical realism of predictions and the model's generalization ability in data-scarce scenarios.

In the domain of aerodynamic coefficient prediction, Transformer-GAN applications remain nascent. Our research demonstrates that such models excel in handling high-dimensional input data, enabling simultaneous prediction of multiple aerodynamic parameters with reduced reliance on data quality and quantity. However, further optimization of model architectures and prediction accuracy remains a critical focus.

The literature review above highlights that while deep learning models have made remarkable progress in airfoil aerodynamic performance prediction, challenges persist. To address these limitations, this study proposes the Deeptrans model, offering a novel, efficient, and accurate solution for aerodynamic performance prediction.

*2.3 Transformer Algorithm*

Since its introduction by Vaswani et al. in 2017, the Transformer model has achieved remarkable success in natural language processing (NLP) and computer vision (CV) due to its powerful sequence modeling capabilities and parallel computation advantages. The core idea of the Transformer lies in its stacked encoder-decoder architecture with multi-head self-attention mechanisms (Dosovitskiy et al., 2020), which capture global dependencies in input sequences. By calculating correlation weights between any two positions in a sequence (as shown in the equation below), the model extracts hierarchical features from different positions and relationships, thereby enhancing its expressive power. After processing through the self-attention mechanism, the extracted features are fed into a feed-forward neural network for nonlinear transformation and further feature extraction. The decoder generates output sequences based on contextual dependencies.

The architecture of the Transformer model is illustrated in the figure below.

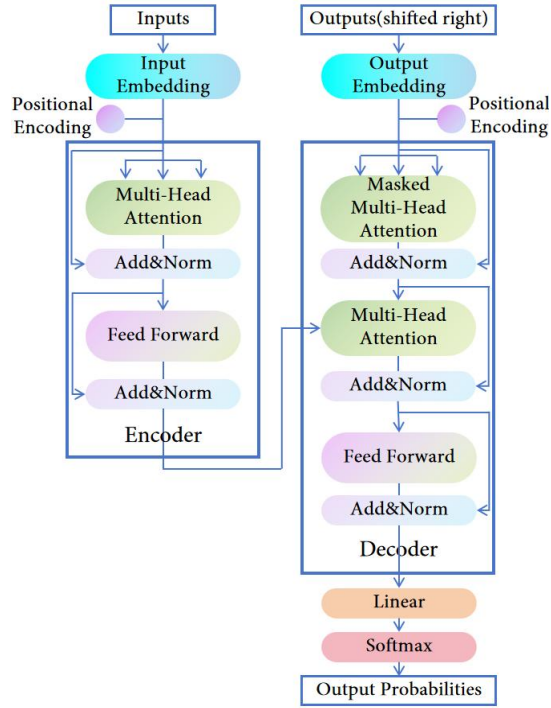

Figure 1.The architecture of the Transformer model.

The correlation weights between any two positions in the sequence are computed as shown in the equation below.

$$\text{Attention}(Q, K, V) = \text{softmax}(\frac{QK^T}{\sqrt{d_k}})V \qquad (1)$$

$Q$, $K$, and $V$ represent the query matrix, key matrix, and value matrix, respectively. $l$ denotes the sequence length, $d$ is the total sequence dimension, and $d_k$ is the dimension of the query matrix $Q$, where $d_k = d/l$.

### 2.4 GAN Algorithm

Generative Adversarial Network (GAN), proposed by Goodfellow et al. in 2014, employs adversarial training between a generator ($G$) and a discriminator ($D$) to model data distributions. Generator $G$ aims to produce synthetic data capable of deceiving the discriminator, while $D$ attempts to distinguish real data from generated data. The two networks optimize their objective function through a minimax game, as formulated below:

$$\min_{G} \max_{D} V(G,D) = \mathcal{E}_{x \sim p_x(x)}[\log D(x)] + \mathcal{E}_{z \sim p_z(x,z)}[\log(1 - D(G(X,Z)))] \tag{2}$$

$p_x(x)$ represents the distribution of real data, and $p_z(x,z)$ denotes the latent space distribution. The core architecture of GAN is illustrated in the figure.

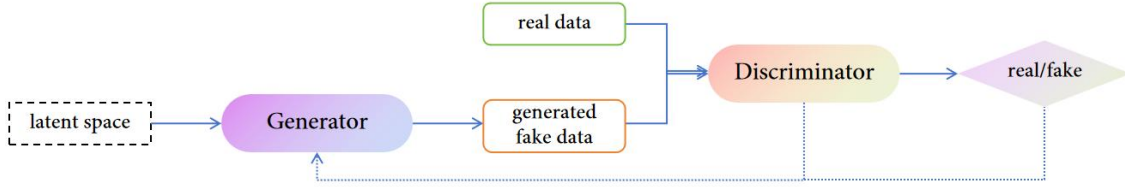

Figure 2. The core architecture of GAN.

The generator network $G(z,\theta)$ maps a noise vector $z$, randomly sampled from the latent space to the real data domain, where $\theta$ denotes the learnable parameters of $G$. The discriminator network $D(x,\phi)$ evaluates the authenticity of input data by outputting a probability score, with $\phi$ representing the parameters of $D$.

Classic GANs often suffer from stability issues such as mode collapse. To address this, Arjovsky et al. proposed the Wasserstein GAN (WGAN) in 2017, which replaces the Jensen-Shannon divergence with the Wasserstein distance to measure the discrepancy between generated and real distributions. Building on this, Gulrajani et al. introduced WGAN-GP, a key improvement involving the addition of a gradient penalty term to the discriminator's loss function. This enforces the 1-Lipschitz continuity condition by constraining the function space $\pi$, thereby enhancing training robustness. The modified objective function is expressed as:

$$\min_{G} \max_{D \in \pi} V(G,D) = \mathcal{E}_{z \sim p_z(x,z)}[D(G(X,Z))] - \mathcal{E}_{x \sim p_x(x)}[D(x)] + \lambda \mathcal{E}_{u \sim p_u(x,u)}[(\|\nabla_\mu D(X,U)\| - 1)^2] \tag{3}$$

This regularization strategy effectively mitigates vanishing or exploding gradients, significantly improving the diversity of generated samples and the convergence efficiency of training.

# 3 Prediction of Aerodynamic Performance Parameters from Airfoil Coordinates Files Based on an Improved Transformer Model

## *3.1 Dataset Construction*

### *3.1.1 Construction of Airfoil Coordinates Dataset*

This study utilizes the University of Illinois UIUC Airfoil Database, which contains 1,477 distinct airfoils. Each airfoil is provided in two formats: Lednicer and Selig. Our database adopts the Lednicer format. Since the number of coordinate points in these files is not fixed, we uniformly sampled 16 points from the upper surface and 15 points from the lower surface, generating a 31×2 matrix to represent the original airfoil geometry. To incorporate environmental conditions, the first row of the matrix includes the Reynolds number (*Re*) and the Ncrit factor ($n_{crit}$). To mitigate training instability caused by large Reynolds number values, we encode them as discrete integers (1~5) corresponding to *Re*=50000,100000,200000,500000 and 1000000.This results in a final 32×2 input matrix.

For example, the cross-sectional contour of the falcon airfoil and its corresponding coordinate matrix under *Re*=200000 and $n_{crit}$=5 are illustrated in the figure and table below, respectively.

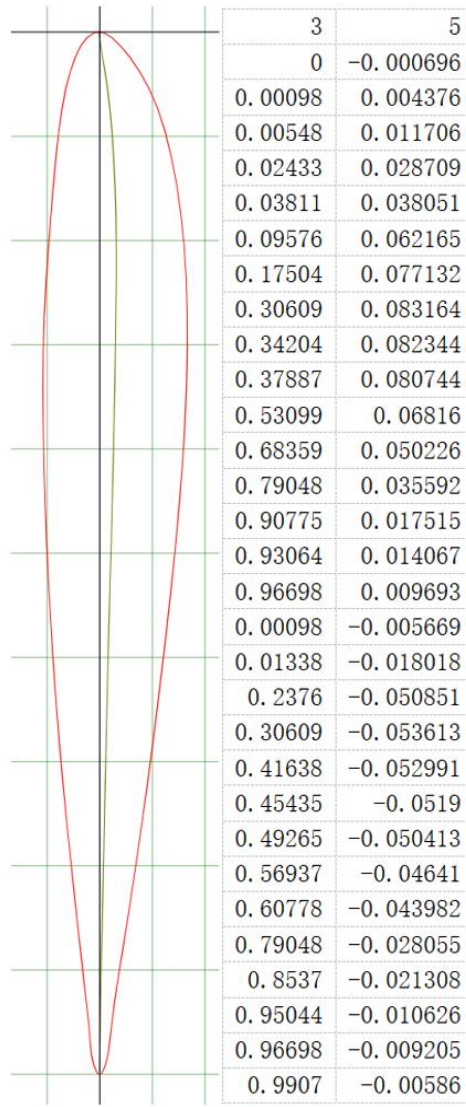

| 3 | 5 |
|---|---|
| 0 | -0.000696 |
| 0.00098 | 0.004376 |
| 0.00548 | 0.011706 |
| 0.02433 | 0.028709 |
| 0.03811 | 0.038051 |
| 0.09576 | 0.062165 |
| 0.17504 | 0.077132 |
| 0.30609 | 0.083164 |
| 0.34204 | 0.082344 |
| 0.37887 | 0.080744 |
| 0.53099 | 0.06816 |
| 0.68359 | 0.050226 |
| 0.79048 | 0.035592 |
| 0.90775 | 0.017515 |
| 0.93064 | 0.014067 |
| 0.96698 | 0.009693 |
| 0.00098 | -0.005669 |
| 0.01338 | -0.018018 |
| 0.2376 | -0.050851 |
| 0.30609 | -0.053613 |
| 0.41638 | -0.052991 |
| 0.45435 | -0.0519 |
| 0.49265 | -0.050413 |
| 0.56937 | -0.04641 |
| 0.60778 | -0.043982 |
| 0.79048 | -0.028055 |
| 0.8537 | -0.021308 |
| 0.95044 | -0.010626 |
| 0.96698 | -0.009205 |
| 0.9907 | -0.00586 |

Figure 3.The cross-sectional contour of the falcon airfoil and its input coordinate matrix under *Re*=200000 and $n_{crit}$=5.

*3.1.2 Construction of Aerodynamic Performance Parameter Dataset*

This study employs Xfoil, an application developed by Mark Drela and Harold Youngren for the design and analysis of subsonic airfoils, to generate aerodynamic performance matrices for various airfoils. Since our research is based on the default computational model assuming incompressible flow, the Mach number is set to 0. The Reynolds number ranges logarithmically from 50,000 to 1,000,000, representing a dimensionless quantity dependent on velocity, airfoil chord length, and fluid properties. The Ncrit factor is used to simulate turbulence effects or surface roughness of the airfoil.

A section in the Xfoil documentation explains this methodology, accompanied by a table of recommended values (shown below):

| Situation | Ncrit |
|:---:|:---:|
| sailplane | 12 to 14 |
| motorglider | 11 to 13 |
| clean wind tunnel | 10 to 12 |
| average wind tunnel | 9 |
| dirty wind tunnel | 4 to 8 |

Table 1.The table of recommended values of the $n_{crit}$ factor under different situations.

To simplify the analysis, two $n_{crit}$ values (5 and 9) were selected. Polar curves were initially generated over an angle of attack ($\alpha$) range of -20° to 20° with a step size of 0.25°. Due to convergence failures at certain points, the final angle range may vary or become discontinuous. Xfoil's lift and drag predictions are considered valid slightly beyond the *maximum lift coefficient* ($C_{L_{max}}$), though this range may extend significantly depending on the airfoil geometry and Reynolds number. After generating polar files, they were post-processed to detect the $C_{L_{max}}$ and $C_{L_{min}}$, with truncation applied beyond these values. To standardize the input matrix size, aerodynamic parameters—lift coefficient ($C_L$), drag coefficient ($C_D$), pressure drag coefficient ($C_{D_p}$), pitching moment coefficient ($C_M$), top transition point (*Top_Xtr*), and bottom transition point (*Bot_Xtr*)——were extracted at angles of -4°,-3°,-2°,-1°,0°,1°,2° and 3° to form an 8×7 matrix. Each airfoil analysis yielded 10 aerodynamic performance tables (5 Reynolds numbers× 2 Ncrit factors).

| Alpha | CL | CD | CDp | CM | Top_Xtr | Bot_Xtr |
|---|---|---|---|---|---|---|
| -4 | -0.2939 | 0.0136 | 0.0054 | -0.0211 | 0.8868 | 0.1377 |
| -3 | -0.1979 | 0.0116 | 0.0041 | -0.0182 | 0.8092 | 0.3484 |
| -2 | -0.1134 | 0.0104 | 0.0036 | -0.0117 | 0.7234 | 0.6122 |
| -1 | -0.0166 | 0.0102 | 0.0035 | -0.0071 | 0.6494 | 0.7643 |
| 0 | 0.0949 | 0.0105 | 0.0036 | -0.0058 | 0.5878 | 0.8635 |
| 1 | 0.2249 | 0.0109 | 0.0037 | -0.0092 | 0.5322 | 0.9137 |
| 2 | 0.3678 | 0.0115 | 0.004 | -0.0159 | 0.4772 | 0.9452 |
| 3 | 0.5113 | 0.0122 | 0.0045 | -0.0229 | 0.4217 | 0.9684 |

Table 2. The aerodynamic performance table of the falcon airfoil under *Re*=200000 and $n_{crit}$=5.

### *3.2 Basic Framework of the Improved Transformer Generation Model*

The improved Transformer-based generative model (Deeptrans) proposed in this study integrates generative adversarial networks (GAN) and leverages core Transformer components to construct an aerodynamic performance prediction system. The model is designed to predict an 8×7 aerodynamic performance matrix from a 32×2 airfoil coordinate matrix. The input matrix comprises 32 time steps of airfoil sequence data:

Row 1: Global features ($n_{crit}$ and *Re*).

Rows 2–17: Upper surface coordinates (*x,y*) of the airfoil.

Rows 18–32: Lower surface coordinates (*x,y*) of the airfoil.

The output matrix corresponds to seven aerodynamic performance coefficients across eight angles of attack.

The overall architecture of the model is illustrated in the figure below.

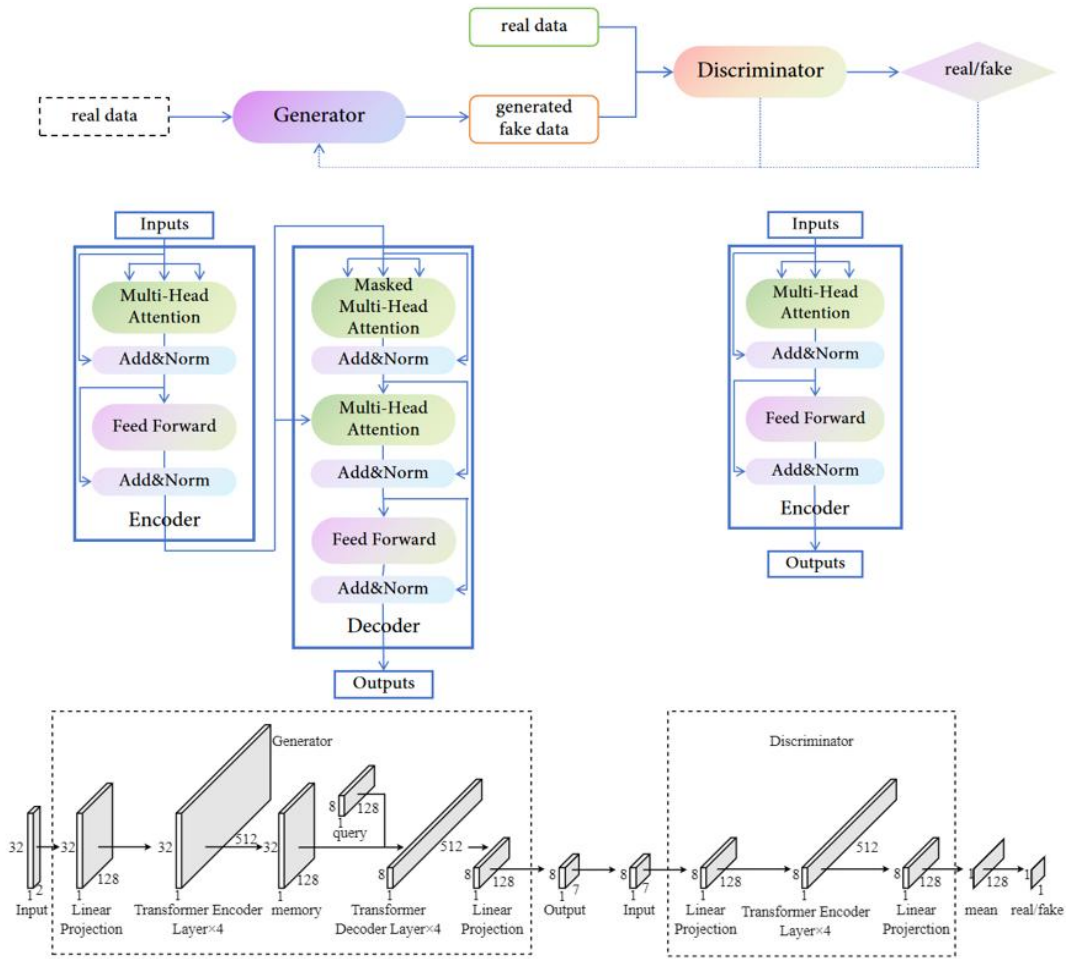

Figure 4.The overall architecture of the Deeptrans model.

### 3.2.1 Generator Network Design

The generator is designed based on a Transformer encoder-decoder framework and includes the following key modules:

*Input Projection Layer:*

A linear layer processes the input matrix (32×2) to map 2D coordinates into a 128-dimensional embedding space, preserving geometric topology relationships for Transformer computation. Experimental results indicate that removing the traditional positional encoding module in the Transformer architecture reduces training loss for this task.

*Transformer Multi-Scale Feature Encoder:*

A 3-layer Transformer encoder with 4-head self-attention mechanisms (nhead=4) captures spatial correlations between airfoil coordinate points. The lower layers focus on local geometric variations (e.g., leading-edge curvature, lower-surface concavity), while higher layers establish cross-regional interactions (e.g., pressure coupling between upper and lower surfaces).

*Angle-of-Attack Adaptive Decoder:*

Eight learnable query vectors are introduced, each corresponding to a specific angle of attack (-4° to 3°). The decoder extracts aerodynamic performance information via cross-attention mechanisms, enabling interaction between queries and encoded memory. This dynamically focuses on critical flow features under varying angles of attack. Orthogonality constraints on the query vectors ensure prediction independence across angles.

*Output Projection Layer:*

A linear projection layer maps the 128-dimensional features to 7 aerodynamic parameters, reconstructing the output sequence into an 8×7 matrix format.

### 3.2.2 Discriminator Network Design

*Input Projection Layer:*

A linear layer projects the 8×7 aerodynamic performance matrix into a 128-dimensional embedding space, providing suitable features for subsequent Transformer modules.

*Transformer Encoder:*

A 4-layer Transformer encoder employs self-attention mechanisms to learn relationships between angles of attack and aerodynamic coefficients, identifying characteristic aerodynamic patterns across different angles.

*Fully Connected Output Layer:*

Mean pooling aggregates features by compressing the temporal dimension, capturing holistic flow-field characteristics. The final layer outputs a single probability value to assess the authenticity of input data.

*3.2.3 Training Mechanism and Optimization Strategy*

*Adversarial Training Mechanism*

Based on the generative adversarial network (GAN) framework, the model achieves efficient training through dynamic competition between the generator and discriminator. The discriminator employs a 3-layer Transformer encoder to extract nonlinear coupling relationships among aerodynamic parameters. It utilizes mean pooling to capture holistic flow-field characteristics and implicitly learns simplified constraints from the Navier-Stokes equations, thereby suppressing unphysical solutions. To stabilize training, the discriminator is updated 5 times ($n\_critic = 5$) for every 1 update of the generator. This design ensures the discriminator converges to a stable state first, preventing the generator from prematurely falling into local optima.

*Improved Huber Loss Function*

The Huber loss replaces the traditional MSE to balance robustness against outliers and convergence speed. The Huber function uses MSE for small errors () to accelerate convergence and switches to MAE for large errors () to suppress outlier interference. During initial training, we observed that the last column of the generated matrix exhibited the largest errors compared to the ground truth. To prioritize this column, we augmented its weight in the loss calculation. Experimental results confirm that this adjustment further reduces the post-training loss value. The modified Huber loss function is formulated as:

$$L_{\text{Huber}} = \sum_{i=1}^{B}\sum_{j=1}^{8}\sum_{k=1}^{7} w_k \cdot \begin{cases} 0.5 \cdot r_{ijk}^2 & \text{if } |r_{ijk}| < \delta, \\ \delta \cdot (|r_{ijk}| - 0.5\delta) & \text{otherwise,} \end{cases}$$

$$w_k = \begin{cases} 2.0 & \text{while } k = 7, \\ 1.0 & \text{other column,} \end{cases} \quad (4)$$

Where B denotes the batch size, $r_{ijk} = recon\_x_{ijk} - x_{ijk}$ represents the residual between the predicted value and the ground truth for the *i*-th sample, *j*-th row, and *k*-th column. The threshold parameter $\delta$ for the Huber loss is set to 1.0. The column weight coefficient $w_k$ is 2.0 when *k*=7 (corresponding to the last column of the output matrix) and 1.0 otherwise.

*Dynamic Optimization Strategy*

A dual Adam optimizer (initial learning rate: $1 \times 10^{-3}$) is employed, combined with a ReduceLROnPlateau scheduler that reduces the learning rate by 50% when validation loss plateaus, enhancing convergence. Gradient clipping (clip_grad_norm=1.0) is applied to prevent gradient explosion. To ensure stable discriminator performance, the generator is updated once after every 5 discriminator training iterations. This strategy prioritizes discriminator convergence and maintains training stability through gradient clipping.

*Training Balance Mechanism*

The total generator loss is defined as $L_G = L_{adv} + L_{Huber}$, where $L_{adv}$ is an adversarial loss that drives the generated data distribution to approximate the real data distribution and $L_{Huber}$ is a reconstruction loss that ensures pointwise accuracy. The discriminator loss is computed as $L_D = (L_{D_{\text{real}}} + L_{D_{\text{fake}}})/2$, which balances losses from real and fake data to prevent the discriminator from overpowering the generator prematurely. All losses are averaged over the batch size to eliminate scaling effects caused by varying batch dimensions, ensuring hyperparameter generalizability.

*3.2.4 Model Training Results*

To achieve optimal performance for the proposed Deeptrans model, we systematically conducted hyperparameter tuning experiments on critical parameters. The tuned hyperparameters included:

*nhead*: Number of heads in the multi-head attention mechanism.

*numlayers*: Number of layers in the Transformer encoder and decoder.

*colweight*: Weighting coefficient for the last column in the Huber loss.

*learning rate*: Shared learning rate for both the generator and discriminator.

*batch size*: Number of samples per batch during forward/backward propagation.

The experiments utilized a dataset comprising 1,438 airfoil files (10,553 training samples and 1,173 validation samples). The model was trained on a single NVIDIA GeForce RTX 3060 Laptop GPU. All non-target hyperparameters were fixed to default values(*nhead* = 4, *numlayers* = 4, *colweight* = 2, *learning rate* = $10^{-3}$, *batch size* = 640). Each configuration was trained for 150 epochs, and the resulting MSE loss and training time are summarized below:

| nhead | 2 | 4 | 8 | 16 |
|---|---|---|---|---|
| Training loss | 0.0000089 | 0.0000054 | 0.0000077 | 0.0000061 |
| Validation loss | 0.0000093 | 0.0000056 | 0.0000058 | 0.0000055 |
| Training time | 941.10s | 1091.88s | 1315.68s | 1595.12s |

| numlayers | 2 | 3 | 4 | 5 |
|---|---|---|---|---|
| Training loss | 0.0000089 | 0.0000064 | 0.0000054 | 0.0000077 |
| Validation loss | 0.0000085 | 0.0000067 | 0.0000056 | 0.0000087 |
| Training time | 559.99s | 820.32s | 1091.88s | 1342.54s |

| colweight | 1 | 2 | 3 |
|---|---|---|---|
| Training loss | 0.0000061 | 0.0000054 | 0.0000063 |
| Validation loss | 0.0000065 | 0.0000056 | 0.0000064 |
| Training time | 559.99s | 1091.88s | 1078.62s |

| learning rate | $10^{-2}$ | $10^{-3}$ | $10^{-4}$ |
|---|---|---|---|
| Training loss | 0.0012791 | 0.0000054 | 0.0000238 |
| Validation loss | 0.0013551 | 0.0000056 | 0.0000230 |
| Training time | 1075.89s | 1091.88s | 1081.86s |

| batch size | 384 | 512 | 640 | 768 | 896 |
|---|---|---|---|---|---|
| Training loss | 0.0000078 | 0.0000080 | 0.0000054 | 0.0000075 | 0.0000059 |
| Validation loss | 0.0000111 | 0.0000106 | 0.0000056 | 0.0000088 | 0.0000081 |
| Training time | 1140.18s | 1115.21s | 1091.88s | 813.46s | 792.44s |

Table 3.Model losses and training time across different hyperparameter configurations.

After comprehensively evaluating both loss metrics and training duration, the optimal configuration for the Deeptrans model was determined as *nhead* = 4, *numlayers* = 4, *colweight* = 2, *learning rate* = $10^{-3}$, and *batch size* = 640.

Using this configuration, we further analyzed the relationship between training loss, validation loss, training time, and dataset size, as summarized in the table below:

| Number of Airfoils | 500 | 1000 | 1438 |
|---|---|---|---|
| Training dataset size | 3700 | 7208 | 10553 |
| Validation dataset size | 412 | 801 | 1173 |
| training loss | 0.0000177 | 0.0000077 | 0.0000054 |
| validation loss | 0.0000236 | 0.0000114 | 0.0000056 |
| training time | 374.69s | 737.92s | 1091.88s |

Table 4.Model losses and training time across different dataset sizes.

The table illustrates that as the training sample size expands, both the training loss and validation loss show significant improvement, indicating the strong generalization capability of the proposed model. When the full dataset of 1,438 airfoil samples is employed, the validation error decreases to $5.6 \times 10^{-6}$, demonstrating the model's robust ability to predict airfoil aerodynamic performance.

**4 Case study**

*4.1 Case Description*

In 2024, the University of Illinois UIUC Airfoil Database introduced several new airfoils. However, aerodynamic performance data for these airfoils are not yet available on http://airfoiltools.com/. To validate the model's performance, four newly added airfoils—cal2263m, ds21, cal1215j, and legionair140_sm—were selected. These airfoils were analyzed under 10 distinct operating conditions, combining Reynolds numbers (*Re*=50,000;100,000;200,000;500,000;1,000,000) and the Ncrit factor ($n_{crit}$=2 or 5), using both Xfoil and the proposed deep learning model. The experiment was conducted on a system equipped with a 12th Gen Intel Core i7-12700H CPU and an NVIDIA GeForce RTX 3060 Laptop GPU.

*4.2 Prediction Results*

*4.2.1 Prediction Efficiency*

The table below compares the time required by Xfoil and the Deeptrans model to compute aerodynamic performance matrices for individual operating conditions (specified *Re* and $n_{crit}$). Experimental results show that Xfoil requires an average of 3.8250 seconds per computation, while Deeptrans achieves the same task in only 0.005567 seconds on average.

| Method | Airfoil | Time |
| --- | --- | --- |
| Xfoil | cal2263m | 1.6996 |
| | ds21 | 8.6199s |
| | cal1215j | 1.3773s |
| | legionair140_sm | 3.6031s |
| Deeptrans | cal2263m | 0.005734s |
| | ds21 | 0.005406s |
| | cal1215j | 0.005882s |
| | legionair140_sm | 0.005244s |

Table 5. Aerodynamic computation time comparison between Xfoil and Deeptrans.

*4.2.2 Prediction Accuracy*

To evaluate the predictive capability of the Deeptrans neural network, we analyzed three representative cases: Airfoil cal2263m at *Re*=500,000 and $n_{crit}$=5, Airfoil ds21 at *Re*=50,000 and $n_{crit}$=5, Airfoil cal1215j at *Re*=1,000,000 and $n_{crit}$=9. The results (illustrated in accompanying figures) demonstrate strong agreement between Xfoil calculations and Deeptrans predictions. Absolute error heatmaps (rightmost panels) visualize discrepancies between Xfoil and Deeptrans matrices. The maximum absolute errors were observed as follows: lift coefficient ($C_L$) reached 0.053, drag coefficient ($C_D$) 0.005, pressure drag coefficient ($C_{D_p}$) 0.003, pitching moment coefficient ($C_M$) 0.005, top transition point (*Top_Xtr*) 0.020 and bottom transition point (*Bot_Xtr*) 0.100.

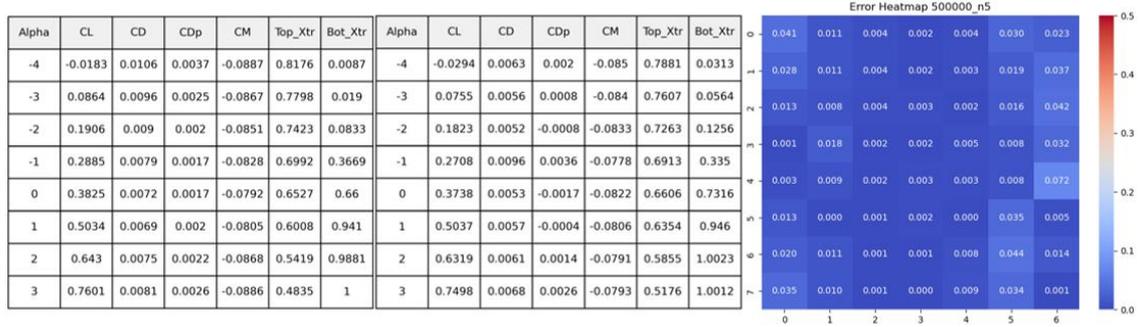

Figure 5. The calculated and predicted aerodynamic performance tables and the absolute error heatmap of the cal2263m airfoil at *Re*=500,000 and $n_{crit}$=5.

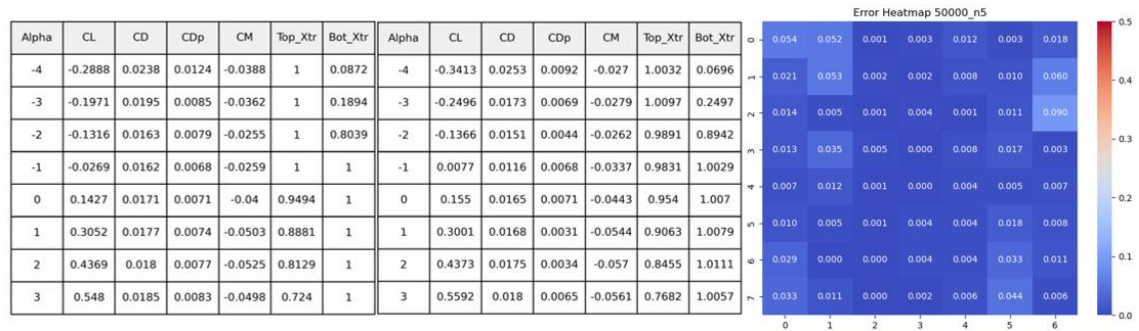

Figure 6. The calculated and predicted aerodynamic performance tables and the absolute error heatmap of the ds21 airfoil at *Re*=50,000 and $n_{crit}$=5.

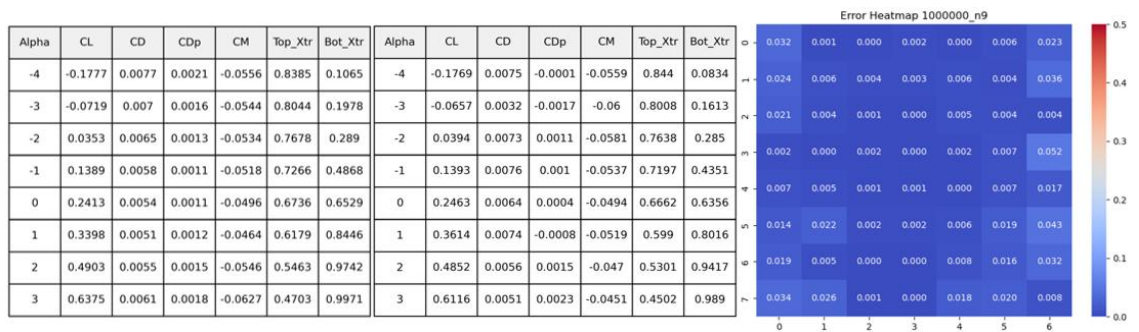

Figure 7. The calculated and predicted aerodynamic performance tables and the absolute error heatmap of the cal1215j airfoil at *Re*=1,000,000 and $n_{crit}$=9.

To further quantify prediction accuracy, we computed Pearson correlation coefficients (*r*) between predicted and actual matrices.

Pearson Correlation Formula:

$$r = \frac{\text{cov}(X,Y)}{\sigma_X \sigma_Y} = \frac{\sum (X_i - \overline{X})(Y_i - \overline{Y})}{\sqrt{\sum (X_i - \overline{X})^2 \sum (Y_i - \overline{Y})^2}} \tag{5}$$

Here, for actual aerodynamic coefficients from Xfoil $X_i \in X$ and predicted aerodynamic coefficients from Deeptrans $Y_i \in Y$, the covariance function is represented by cov(·), and $\sigma_X$, $\sigma_Y$ are standard deviations of $X$ and $Y$. $X$ embodies the actual values of aerodynamic coefficients, while $Y$ embodies the corresponding predicted values. Moreover, $\overline{X}$ and $\overline{Y}$ symbolize respectively the mean values of $X$ and $Y$.

For the legionair140_sm airfoil, the correlation coefficients for $C_L$, $C_D$, $C_{D_p}$, $C_M$, *Top_Xtr,* and *Bot_Xtr* are 1.00, 0.92, 0.96, 0.85, 0.98, and 0.99, respectively. These near-perfect correlations, evident in the plotted data comparisons, validate the high fidelity of Deeptrans predictions.

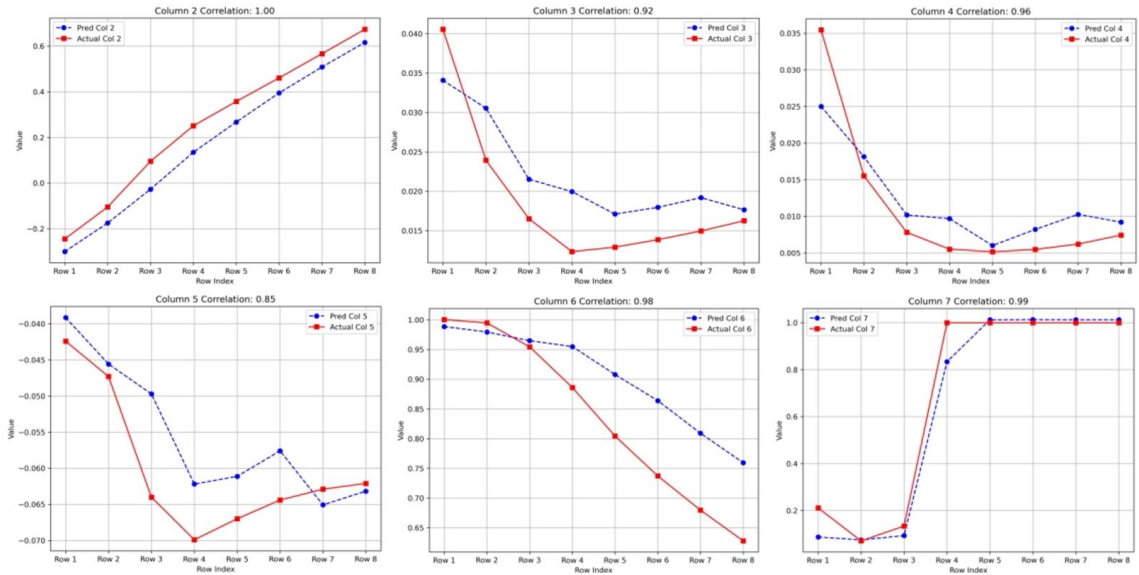

Figure 8.The Pearson correlation coefficient curves for $C_L$, $C_D$, $C_{D_p}$, $C_M$, *Top_Xtr*, and *Bot_Xtr* for the legionair140_sm airfoil.

*4.3 Comparative Analysis*

To further demonstrate the superiority of the Deeptrans model in predicting airfoil aerodynamic coefficient matrices, we trained three additional deep learning models—Transformer, VAE, and GAN—on the same airfoil dataset. The training configurations, including dataset size, batch size, epoch count, final MSE loss, and training time, are summarized in the table below:

|  | Deeptrans | Transformer | VAE | GAN |
|---|---|---|---|---|
| Number of Airfoils | 1438 | 1438 | 1438 | 1438 |
| Training dataset size | 10553 | 10553 | 10553 | 10553 |
| Validation dataset size | 1173 | 1173 | 1173 | 1173 |
| Batch size | 640 | 640 | 640 | 640 |
| Epoch | 150 | 150 | 150 | 150 |
| Training loss | 0.0000054 | 0.0000160 | 0.0000168 | 0.0000223 |
| Validation loss | 0.0000056 | 0.0000182 | 0.0000205 | 0.0000245 |
| Training time | 1091.88s | 36.34s | 63.93s | 27.86s |

Table 6. Model performance comparison between Deeptrans and other baseline models.

The training and testing of all four projects were conducted on a single NVIDIA GeForce RTX 3060 Laptop GPU. During the training process for the four models, the training set comprised 10,553 samples, and the validation set included 1,173 samples. All models were trained using the Adam optimizer and ReduceLROnPlateau learning rate scheduler to ensure optimal convergence. During the initial phase, Transformer, VAE, and GAN were trained with loss functions tailored to their architectures. For fair comparison, their errors were quantified using the same MSE loss function:

$$L_{MSE} = \frac{1}{N} \sum_{i=1}^{N} (recon\_x_i - x_i)^2 \qquad (6)$$

Where $recon\_x_i$ is model-reconstructed output, $x_i$ is ground truth, $N$ is the total number of elements in the input tensor ($N = 8 \times 7 = 56$).

The training loss curves for each model are shown in the accompanying figures.

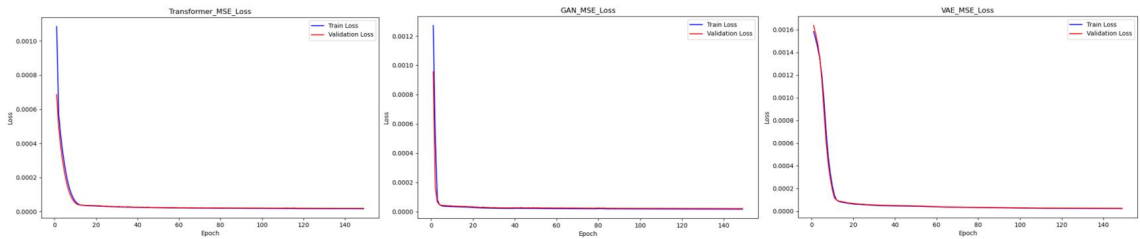

Figure 9.The training loss curves of Transformer, GAN, and VAE models.

After 150 epochs, the validation losses for the Transformer, VAE, and GAN were 0.0000182, 0.0000205, and 0.0000245, respectively. To address these limitations, the Transformer and GAN architectures—exhibiting the lowest losses among the three—were integrated and refined to develop the Deeptrans model. Retaining the same optimizer and scheduler, Deeptrans was trained for 150 epochs on the identical dataset. The validation loss of Deeptrans decreased to 0.0000056 (achieved in 1,091.88 seconds), significantly outperforming the baseline models. The loss curve for Deeptrans is illustrated in the figure below.

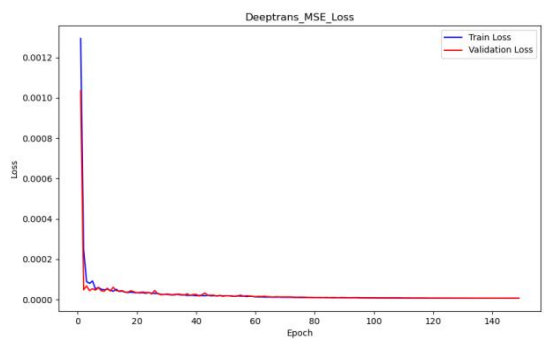

Figure 10.The training loss curve of the Deeptrans model.

**5 Discussion and conclusion**

In this study, we propose Deeptrans, a deep learning model combining the Transformer encoder architecture with principles from generative adversarial networks

(GAN), for predicting aerodynamic performance matrices of airfoils under specific flow conditions. The model was trained and evaluated using a large-scale dataset comprising 10,553 training samples and 1,173 validation samples. Through systematic comparative experiments, we analyzed the impact of multiple hyperparameters on model performance and identified the optimal configuration. The final model demonstrates outstanding accuracy, efficiency, and generalization capability when trained on the complete dataset.

## 5.1 Innovation and Contribution

The core innovation and contribution of this work can be summarized as the following three points:

Firstly, the proposed Deeptrans model demonstrates exceptional computational efficiency in single-sample aerodynamic performance prediction. With a prediction time of 0.0056 seconds per case on average, it achieves a 687-fold acceleration compared to conventional CFD methods (benchmarked against the Xfoil method averaging 3.825 seconds). This advancement significantly enhances the computational speed for obtaining aerodynamic parameters of airfoils, providing an efficient approach for rapid approximation of aerodynamic performance in engineering applications.

Second, we constructed a standardized aerodynamic dataset for airfoil performance prediction. Starting from raw airfoil coordinate points and high-fidelity CFD-derived aerodynamic parameters, we established a unified parametric framework integrating both $n_{crit}$ and $Re$. The dataset consists of 1438 rigorously validated sample airfoil types. It provides a standardized, scalable, and high-quality training resource, establishing a reliable foundation for advancing deep learning applications in airfoil prediction tasks.

Finally, we propose Deeptrans, a hybrid Transformer-GAN architecture that synergizes global contextual modeling with physics-aware adversarial refinement. By integrating the long-range dependency modeling strengths of Transformers with the data generation quality enhancement capabilities of GANs, Deeptrans achieves significantly improved prediction accuracy. Experimental results confirm that Deeptrans outperforms traditional fully connected networks, CNNs, and standalone Transformers under identical data conditions, validating its superiority in aerodynamic performance prediction.

*5.2 Limitations and Future Work*

Although this study has achieved certain results, the following limitations remain and warrant further exploration and improvement in future work:

*Dataset Quality and Uniformity Require Enhancement*

Current discrepancies in the number of airfoil coordinate points and the range of angle-of-attack parameters necessitated uniformity constraints. Specifically, only 31 coordinate points and 8 predefined angles of attack were sampled for training. This approach partially limits the model's ability to comprehend geometric diversity and restricts its applicability. Future efforts should expand the dataset scale using high-fidelity CFD data to improve diversity and quality.

*Training and Prediction Efficiency Can Be Further Optimized*

While Deeptrans achieves a significant speed advantage in single-sample prediction, the overall training time remains constrained by hardware and code efficiency. Subsequent work should explore strategies such as model pruning, distributed training, or hardware upgrades to accelerate the training process.

*Lack of Broad Flow Parameter Integration*

The current model does not incorporate Mach number or other extended flow parameters, limiting its adaptability to high-speed aerodynamic scenarios. Future research should investigate alternative simulation methods under high Mach number conditions and develop an extended version of Deeptrans with Mach number inputs to enhance adaptability and application scope.

*Physical Consistency and Interpretability Require Enhancement*

The current model is primarily trained using a data-driven approach, lacking physical constraint mechanisms. In future research, we plan to integrate physical priors (e.g., momentum conservation, lift-drag relationships) as model constraints, developing physics-guided deep neural networks for flow field prediction. This will improve the model's interpretability and credibility.

In summary, the proposed Deeptrans model demonstrates promising performance and application prospects in airfoil aerodynamic performance prediction. Future work will focus on optimizing the model architecture and data framework, incorporating additional flight parameter dimensions (e.g., Mach number), and advancing the practical application of deep learning methods in complex aerodynamic prediction challenges.

**Disclosure statement**

No potential conflict of interest was reported by the authors.

**Funding**

This work was supported by the National Natural Science Foundation of China (No. 52305280) and the Natural Science Basic Research Program of Shaanxi Province (No. 2023-JC-QN-0397).